\documentclass[10pt,twocolumn,letterpaper]{article}

\usepackage{wacv}              

\usepackage{graphicx}
\usepackage{amsmath}
\usepackage{amssymb}
\usepackage{booktabs}
\usepackage{enumitem}
\setlist[itemize]{leftmargin=3mm}

\usepackage[pagebackref,breaklinks,colorlinks]{hyperref}

\usepackage[capitalize]{cleveref}
\crefname{section}{Sec.}{Secs.}
\Crefname{section}{Section}{Sections}
\Crefname{table}{Table}{Tables}
\crefname{table}{Tab.}{Tabs.}

\def\wacvPaperID{2703} 
\def\confName{WACV}
\def\confYear{2025}

\begin{document}

\title{Using Prompts to Guide Large Language Models in Imitating a Real Person's Language Style}

\author{
Ziyang Chen \\
\and
Stylios Moscholios
}
\maketitle

\begin{abstract}
Large language models (LLMs), such as GPT series and Llama series have demonstrated strong capabilities in natural language processing, contextual understanding, and text generation. In recent years, researchers are trying to enhance the abilities of LLMs in performing various tasks, and numerous studies have proved that well-designed prompts can significantly improve the performance of LLMs on these tasks. This study compares the language style imitation ability of three different large language models under the guidance of the same zero-shot prompt. It also involves comparing the imitation ability of the same large language model when guided by three different prompts individually. Additionally, by applying a Tree-of-Thoughts (ToT) Prompting method to Llama 3, a conversational AI with the language style of a real person was created. In this study, three evaluation methods were used to evaluate LLMs and prompts. The results show that Llama 3 performs best at imitating language styles, and that the ToT prompting method is the most effective to guide it in imitating language styles. Using a ToT framework, Llama 3 was guided to interact with users in the language style of a specific individual without altering its core parameters, thereby creating a text- based conversational AI that reflects the language style of the individual.

\end{abstract}

\section{INTRODUCTION}

Conversational AI, such as chatbots based on large language models (LLMs), has been widely implemented as virtual assistants in various domains of daily life, including medical and health consulting, social entertainment, and personal scheduling. With the advancement of artificial intelligence technology, the language style of conversational AI is gradually becoming more personalized, increasingly approaching the authentic language style of real humans. Conversational AI is a key component of digital human technology, as it enables digital humans to effectively communicate with human beings. Moreover, conversational AI can be integrated with AI cloning technology, allowing digital humans to imitate the language styles of real individuals to communicate with humans. One of the main challenges of creating a digital human that uses the language style of a real person is how to accurately emulate the style. To solve this problem, the first step is to enable the key component, conversational AI, imitate the person’s language style. Traditional methods mainly focus on using a large volume of the person’s text to train models. But this study explored the possibility of developing a text-based conversational AI with real person’s language style using only prompt engineering techniques and a limited amount of text of that person.

Prompt engineering is an emerging field that aims to explore the potential of large language models by guiding them to complete specific tasks using prompts composed of natural language. Based on the deep understanding of natural language by large language models, prompt engineering allows the models to be fine-tuned using only prompts and small batches of samples, without requiring extensive data training, thereby reducing training costs. Despite the potential of prompt engineering, there are few studies focusing on how it can be used to help LLMs imitate language styles.

This study aims to further the research on imitation ability of LLMs under prompts, containing three steps. The first step involves comparing the performance of three LLMs when imitating language styles, using the same zero-shot prompt. The second step is to compare the impact of different prompts on the imitation ability of the same large language model. The third step is to use prompts to develop a text-based conversational AI, which keeps the language style of a real person. To conduct the research, three datasets were collected from public interviews with celebrities, and three LLMs, GPT-4, Llama 3, and Gemini 1.5, were selected for comparison. Each model has the basic functions of text generation, text style analysis, and text style transfer. Additionally, three different types of promptings, Zero-Shot Prompting, Chain-of-Thought (CoT) prompting, and Tree-of- Thoughts (ToT) prompting, were compared by applying them to a same large language model, respectively.

To evaluate the imitation ability of different large language models, a zero-shot prompt was designed, which guides these LLMs in imitating the language style of specified individuals based on their texts. The models were asked to generate conversations imitating the specified individual's language style, which would reflect their imitation ability. Through the comparison of different models’ conversations, their imitation abilities can be evaluated. To compare different promptings, the study designed and applied three different prompts to individually guide Llama 3 in imitating the language style of a target individual. For evaluation, three evaluation methods were developed to assess the imitation abilities of the large language models: human evaluation, LLM evaluation, and automated evaluation. To develop the conversational AI, a ToT method was designed, which guides Llama 3 to speak like a real person. It takes the paragraphs from the conversation of a real person, and learns the person’s special language style. As a result, it can chat with the user in the form of textual dialogue, holding the learned language style.

This study compares the ability of different large language models in imitating the language styles of real persons, given dialogue-format text and a zero-shot prompt. It also evaluates how well a large language model can perform the same imitation task when given different prompts. Moreover, a conversational AI based on Llama 3 was designed, which interacts with users in the style of a real person.

\section{BACKGROUND AND RELATED WORK}

Current research on LLMs imitating language styles mainly focuses on the field of text style transfer. Text style transfer is a natural language processing technique, which transfers the style of the original text but preserves its original semantic content. \cite{hu2022text} employed a phrase-based machine translation technique to train a model using Shakespeare’s texts. The trained model can transform the language style of other texts into a Shakespearean language style. \cite{jhamtani2017shakespearizing} used an end-to-end neural network combined with network pointer techniques to transform texts into a Shakespearean style, successfully achieving the style imitation. Traditional methods achieve the goal of imitating language styles by training models on a large amount of data, which leads to high costs. Therefore, an effective alternative is to use prompt engineering techniques to guide large language models in imitating specific language styles through natural language instructions.

Prompt engineering techniques can effectively enhance the ability of large language models to perform tasks. Under the guidance of well-designed prompts, models are able to achieve better performance without modifying their core parameters. There are a variety of prompting methods, such as Zero-Shot Prompting, Few-Shot Prompting, Chain-of- Thought Prompting, and Tree-of-Thoughts prompting. \cite{sahoo2024systematic} summarized and analyzed 29 existing prompt engineering techniques, providing a comprehensive classification and comparison of their respective advantages and disadvantages. In the field of controllable text generation, prompt-based methods can be applied during the fine-tuning phase of large language models, which helps the model to generate text that meets the requirements of the original task \cite{zhang2023survey}.

Research on using prompt engineering techniques to guide large language models in imitating language styles is still in its early stages. \cite{li2023chatharuhi} constructed a dataset containing dialogues from 32 fictional characters in film and television. They used this dataset to fine-tune large language models such as ChatGLM2. Based on prompts of specific characters, the fine-tuned model can play the specified character and imitate its language style during the interaction with users. Liu and Ni (2023) constructed a text dataset of fictional characters from novels, and used prompts to guide LLMs in setting up character scenarios and background information. Under the guidance of well-designed character prompts, the fine-tuned LLMs can answer questions in the specified character's language style and habit.  \cite{bhandarkar2024emulating} conducted the first feasibility study on using promptings to guide pre-trained large language models for controllable stylized text generation. They apply a task-driven prompt to various LLMs, letting them generate texts in the style of the given texts. They also evaluated the generated texts from the LLMs to test the their ability of imitating writing styles.

\section{Methodology}

\subsection{Dataset}

Three datasets were collected for this research, and each of them consists of a conversation between two individuals including their names and the corresponding paragraphs. All datasets were collected from public interviews of celebrities. Dataset 1 involves a conversation between Elon Musk and an interviewer, while Dataset 3 contains a conversation from a different interview of Elon Musk with another interviewer \cite{ted2022elon,fridman2023elon,kollias2024sam2clip2sam,zafeiriou1,Tailor,springer,kollias20247th,kollias2023multi,kollias2024distribution,psaroudakis2022mixaugment,kollias2021distribution,kollias2019face,kollias2019deep,kollias2019expression,kollias2022abaw,kollias2023abaww,kollias2023btdnet,kollias2023facernet,kollias2023multi,kollias20246th,kollias2024distribution,kolliasijcv,zafeiriou2017aff,hu2024bridging,psaroudakis2022mixaugment,kollias2020analysing,kollias2021distribution,kollias2021affect,kollias2019face,kollias2021analysing,kollias2023ai,kollias2023deep2,arsenos2023data,salpea2022medical,arsenos2024uncertainty,karampinis2024ensuring,arsenos4674579nefeli,miah2024can,arsenos2024commonn,cis,hu2024rethinking,gerogiannis2024covid,kollias2024behaviour4all,kollias2024domain}. Additionally, Dataset 2 includes a conversation from an interview of Tom Holland with another interviewer (Shetty, 2023).

To collect the three datasets, I first used OpenAI's Whisper to convert the video to text. I then manually assigned each paragraph to its corresponding author by comparing it with the video and corrected mistakes generated during the transcription process.

\subsection{Pre-processing}

Firstly, since Elon Musk, Tom Holland, and the interviewers are well-known individuals, to prevent any potential influence of their information in the large language model's database on the experiment, ‘Elon Musk’ in Dataset 1 was replaced with ‘Mark1’, ‘Tom Holland’ in Dataset 2 was replaced with ‘Tony’, and ‘Elon Musk’ in Dataset 3 was replaced with ‘Mark2’. Similarly, the names of the interviewers are replaced by ‘Host1’, ‘Host2’ and ‘Host3’, respectively. The information of the three pre-processed datasets are shown in Table I.

Both Dataset 1 and Dataset 2 were divided into a training set containing 50\% of the paragraphs and a test set containing the remaining 50\%. For Dataset 3, 70\% of the paragraphs were used to form the training set, while the remaining 30\% were used for the test set.

The training sets were provided to the LLMs to learn the language styles, while the test sets were compared with the texts generated by LLMs to evaluate their ability in imitating language styles. Datasets 1 and Dataset 2 were used to evaluate the ability of different LLMs to imitate the language style of a target role under the same prompt, while Dataset 3 was used to evaluate the impact of different prompts on a large language model.

\subsection{Models}

In this project, three widely used LLMs were selected for study: GPT-4, Gemini 1.5 and Llama 3. Information of these models is shown in Table II.

\subsection{Task-setting}
This study consists of three steps. The first step is to compare the ability of three LLMs to imitate language styles under the same prompt. The second step is to compare three prompts in terms of their impact on guiding the same LLM to imitate language styles. The last step is to use a prompting method to guide an LLM in interacting with users in the style of a real person. For each step, a specific task was defined to conduct the research.

To compare the LLMs, Task 1 requires having the LLMs generate a new conversation between two new roles, ‘A’ and ‘B’, on any topic, using the language style of the target role from the training set. The new conversation should include 10 paragraphs. Datasets 1 and 2 are used for this task. In Task 1, the two target roles are Mark1 from Dataset 1 and Tony from Dataset 2. Each LLM is guided to separately imitate each of the two target roles.

To compare different promptings, Task 2 involves having the Llama 3 model generate a continuation of the given text as a conversation between two new roles, ‘A’ and ‘B’, using the language style of the target role. The continuation should consist of 10 paragraphs. Dataset 3 is used for this task, and the target role to be imitated in this task is Mark2.
To design a conversational AI that mimics a real person’s language style, Task 3 is to guide the LLM to learn the language style from the given text and to use this style for interacting with humans.

\subsection{Promptings}

1) Prompting for Task 1:
Zero-Shot Prompting is a natural language processing technique that provides a natural language task description to directly guide the model in completing the task without providing input-output mapping as examples \cite{xu2012paraphrasing}.

Bhandarkar et al. (2024) used a 4-element zero-shot prompt to guide LLMs to generate writings in the style of the given authors’ blogs. Inspired by their work, I designed a similar zero-shot prompt to guide the LLMs for Task 1, which also contains 4 elements. The function of each element is shown in Table III, and the full prompt is shown in Table IV.

2) Promptings for Task 2:
To conduct Task 2, I designed three different prompts: a zero-shot prompt, a chain-of-thought prompt, and a tree-of- thoughts prompting framework. Each of these prompts will be applied to Llama 3.

The zero-shot prompt for Task 2 is similar to that used for Task 1, which also contains the same 4 elements, but the target role to be imitated was set as Mark2. In this context, the zero-shot prompt for Task 2 is referred to as ‘standard prompting’, as the following two prompts are based on this standard prompt. The standard prompt is shown in Fig. 1.

Chain-of-Thought Prompting is a method that guides the model to decompose complex tasks into multiple intermediate steps, leading the model to perform step-by-step reasoning on the task \cite{wei2022chain}. In this study, a chain- of-thought prompt was designed for Task 2. This prompt is based on the standard prompt, with the same content in each element, except for the instruction element. In this element, a plan is provided, which guides the LLM to execute the task following three steps. The chain-of-thought prompt is shown in Fig. 1.

Tree-of-Thoughts (ToT) is a framework that guides a language model in exploring multiple reasoning paths during problem-solving. It employs a self-evaluation method at each stage of reasoning to select the optimal choice. Therefore, the model find a best solution to solve problems by always selecting the best choice at each stage. \cite{yao2023tree} used the ToT framework to guide GPT-4 in creative writing tasks. In their experiment, they divided the task into two steps. The first step was to use a prompt to guide the model in generating five plans for the task, and then to select the best one among those plans. The second step involved guiding the model to generate five outputs based on the best plan, and then selecting the best output. The evaluation of both steps was carried by the model under a zero-shot prompt.

Inspired by their works, a ToT prompting method was designed for Task 2. It’s a framework that contains 2 steps, and four prompts were designed for this framework. The first step is to guide the model to generate three plans for imitating language style of the target role. The plans generated then are compared by the model, and the best plan is selected. In this step, a plan prompt was designed, which guides the model in generating plans. The second step is to guide the model in generating three conversations based on the best plan. After comparing different conversations, the one with the style most similar to that of the target role is selected as the final output. In this step, a conversation prompt was designed to guide the model in generating conversations. To evaluate the outputs and select the best output in each step, two vote prompts were designed to evaluate the outputs for each step, respectively. The first vote prompt asked the model to vote for the best plan based on the task description, while the second vote prompt asked the model to vote for the best conversation that most closely matches the language style of the target role. For each step, the model was asked to vote for the best choice five times. Each time, one choice was selected as the best and received a vote. As a result, the choice with the most votes was selected as the best choice. The process of this framework is shown in Fig. 2, and the prompts involved are shown in the Appendix.

3) Promptings for Task 3
To conduct Task 3, another ToT framework was designed to guide the model in interacting with users in the language style of a real person. This framework is composed of two steps. The first step is to generate different versions of descriptions of the person’s language style, and to choose the best description. The second step is to generate three versions of responses to the user’s input based on the analysis chosen, and to choose the one with the language style most similar to that of the person. During the process, four prompts were designed to guide the model in generating descriptions and responses, and selecting the best choices. The prompts are shown in the Appendix.

\subsection{Implementation}

This study focuses on comparing different LLMs under the same prompt, evaluating the impact of different types of prompts on the same model, and designing a conversational AI in the language style of a real person. Therefore, model training is not required. In this context, I used the API of each model to interact with the LLMs and provided them with prompts to conduct the experiments.

1) Implementation for Task 1:
In Task 1, the target roles to be imitated are Tony from Dataset 1 and Mark from Dataset 2. Each of the three LLMs was asked to separately imitating two target roles. This resulted in a total of six role-model pairs. For each role, three different LLMs were tested individually to assess their ability to imitate the role’s language style. This was done by providing them with the zero-shot prompt designed for Task 1 and the training set of the corresponding dataset. The process was repeated 10 times for each role-model pair, and 10 conversations were collected for each role-model pair.
In the Evaluation section, these conversations will be compared to those of the target roles in the test sets to evaluate the model’s imitation ability.

2) Implementation for Task 2:
In Task 2, the target role is Mark2 from Dataset 3. For this task, one of the models being compared, Llama 3, was chosen to perform different prompts as Llama 3 performs the best in Task 1.
To test different prompts, for standard prompt and CoT prompt, a Python script was developed to interact with Llama 3 via API. Each of the two prompts mentioned above was provided to the model with part of the training set of Dataset 3. The maximum content length of Llama 3 is 8,000 tokens, but the training set contains 13,000 words, which means that the model can’t process the whole training set at once. To solve this problem and to make use of the training set, I employed the sliding window technique to divide the training set into multiple text segments. The window size was set to 4,400 words, with a stride of 2,200 words. As a result, the training set was divided into five segments, and each segment contains 4,400 words. Starting from the second segment, each segment contains the last 50\% of the content from the previous segment, thereby improving the effectiveness of the training set.

Once a prompt is selected, the program will place each text segment into the ‘Given Text’ part of the prompt and send the entire prompt to the model. After receiving a response to the prompt, the program will replace the ‘Given Text’ part with the next text segment and generate a new conversation. The program repeats this process until all segments have been sent to the model. In the end, five conversations were generated by the model under this prompt, with each conversation corresponding to a different text segment.

For ToT prompting, two Python scripts were developed to build this framework and conduct this task, which also applied the sliding window technique to split the training set into five text segments and dealt with each segments in sequence. The first script was about generating plans and choosing the best plan for each text segment. First, three plans on how to generate a continuation with the language style of Mark2 for a segment were generated. Then the model was asked to analyze each plan and vote for the best one five times. Finally, the plan with the most votes was returned as the best plan for this text segment. This process was repeated until each text segment had its own best plan. The second script took a list of the best plans of text segments, and generated three conversations for each text segment based on the corresponding best plan. Similar to the evaluation of plans, the model was asked to analyze conversations of each text segment and vote for the one that best matches Mark2's language style. This program finally returned a list of best conversations for text segments. After finishing the implementation of Task 2, five conversations were generated under each prompt.

3) Implementation for Task 3:
To conduct this task, a script was developed to create a conversational AI with the language style of a real person, based on Llama 3. This program used API of Llama 3 to create a chatbot to interact with users. In order to make the chatbot speaks like a real person, a ToT framework was designed in this program. Before the chat began, the first step was to let the model generate three versions of language style descriptions for the given text. The descriptions should cover various aspects of language style as detailed as possible, listing the most frequently used words and phrases. Then the model was asked to choose the one that best describes the language style of the given text. The second step was to generate three responses every time the model receive a user input, and to choose the best response, which best imitate the person’s style. When chatting with the chatbot, the intermediate generations were invisible to the user, and the user only saw the best conversation every time the user input something. Users could communicate with this chatbot by executing the script in the terminal.

\subsection{ EVALUATION}

\subsubsection{ Human evaluation}

In this study, three evaluators were invited to conduct a manual evaluation for both Task 1 and Task 2 to evaluate the similarity between the conversations generated and the corresponding test sets in terms of language style. The evaluators either major in English or are engaged in English- related work, and they demonstrate a profound understanding of English language, making them well-qualified to conduct the human evaluation.

\cite{hu2022text} summarized text style from a linguistic perspective as being formed through the combined influence of four elements: word choice, sentence structure, figurative language, and sentence arrangement. Inspired by this study, a method similar to the human evaluation approach used in a text style transfer task \cite{li2018delete} was designed. When evaluating a conversation, evaluators were asked to rate this conversation on 4 criteria on a Likert scale from 1 to 5: word choice, sentence structure, figurative language, and sentence arrangement. Therefore, the maximum score is 20 points. The scoring was based on the degree of similarity between the conversation and the paragraphs of the target role in the test set.

1) Evaluation for Task1:
When comparing different models in imitating language styles under the same prompt, 10 conversations were collected for each role-model pair. Thus, for each target role, there are 30 conversations from three LLMs. These 30 conversations have been compiled into one file, and each conversation has been assigned with a number from 1 to 30 to distinguish each other. When evaluating conversations of a target role, the same file containing 30 conversations was assigned to each evaluator. They also received the corresponding test set for comparison. Then, evaluators were asked to rate each conversation based on the criteria mentioned above and return the score for each conversation. Finally, for each role-model pair, 30 scores were collected, which reflect the ability of the model in imitating the language style of the target role. For each role-model pair, an average score of the 30 scores was calculated as the final score of the model. Thus, each model received two final scores. The first score was for imitating the first target role, and the second score was for imitating the second target role. The results of human evaluation for Task 1 are shown in Fig. 3.

2) Evaluation for Task 2:
When comparing the impact of different prompts on Llama 3, five conversations were collected for each prompt. Thus, a total of 15 conversations were generated across the three prompts. Similar to the human evaluation of Task 1, these conversations were compiled into one file, and each conversation was assigned with a number from 1 to 15 for identification. The same file was given to each evaluator to be compared with the test set of dataset 3. After scoring based on the criteria, 15 scores were collected for each prompt, and the average score was calculated as the final score for each prompt. Therefore, each prompt received a final score, which reflects how the model's ability to imitate the language style of the target role is influenced by this prompt. The results of human evaluation for Task 2 are shown in Fig. 4.

\subsubsection{LLM Evaluation}

Due to their strong understanding of natural language, large language models are increasingly being used for text evaluation. Yao et al. (2023) designed a zero-shot prompt to guide GPT-4 in evaluating texts for a creative writing task, aiming to assess the impact of the ToT framework on LLMs. \cite{hung2023who} designed a CoT prompt to evaluate the similarity of text styles between two texts for the purpose of author verification. This author verification method, which is fully based on a well-designed prompt, significantly reduces the cost of model training. Inspired by their works, an evaluation prompt was designed to guide another LLM, Claude 3.5, in evaluating the conversations generated by LLMs. This prompt first asks the model to analyze the similarity between the given conversation and the target role’s paragraphs from the corresponding test set in terms of language style. The analysis covers various aspects of language style. Then, the model gives a score to the conversation from 1 to 10 based on its analysis. The evaluation Prompt is shown in the Appendix.

1) Evaluation for Task 1:
When comparing different LLMs, each conversation generated by each LLM was evaluated by Claude 3.5 using the evaluation prompt three times, and the model returned a score for the conversation each time. Finally, for each role- model pair, 30 scores were collected, and the average score of the 30 scores was calculated as the final score, which reflects the model’s ability to imitate the language style of that target role. In Task 1, since there are two target roles, one from Dataset 1 and one from Dataset 2, two final scores were collected for each LLM. The first score is for imitating Mark1 from Dataset 1, and the second score is for imitating Tony from Dataset 2. The results of LLM evaluation for Task 1 are shown in Fig. 3.

2) Evaluation for Task 2
When comparing different prompts, each conversation generated under each prompt was sent to Claude 3.5 along with the evaluation prompt 3 times, and each conversation got 3 scores in total from the model. Finally, for each prompt, 15 scores were collected, and an average score was calculated as the final score to reflect the impact of this prompt on Llama 3’s imitation ability. The results of this evaluation for Task 2 are shown in Fig. 4.

\subsubsection{ Automatic Evaluation}

Bhandarkar et al. (2024) used Authorship Attribution (AA) methods to evaluate the ability of LLMs in imitating writing styles. In their study, AA methods were used to identify whether a text generated by a LLM belonged to a certain author. If a text was predicted to belong to a certain author, it was considered to have a high degree of similarity with the author’s original writings in terms of writing style. \cite{fabien2020bertaa} developed an AA method based on the BERT model. They fine-tuned the BERT model to perform AA tasks and found that it had the best performance when dealing with short texts. Inspired by their works, I trained a binary classification model based on the BERT model for each target role. It was used to determine whether each paragraph of the generated conversations belonged to the target role. If a paragraph was predicted as from the target role, it was considered to have successfully imitated the language style of the target role. Among the three models in comparison, the model with more paragraphs predicted to belong to the target role is considered to perform better at imitating the language style. Therefore, the number of paragraphs predicted as from the target role reflects the imitation ability of the LLMs. Similarly, the prompts can be evaluated by comparing the number of paragraphs predicted to belong to the target role for each prompt.

1) Evaluation for Task 1
To conduct this evaluation, a new dataset was formed by combining paragraphs from the three datasets, which was then used to train the BERT model. This new dataset contains paragraphs from five roles across three datasets, except for Mark2's paragraphs from Dataset 3. Each role contributes 100 paragraphs to the new dataset, which helps the BERT model distinguish the different roles' language styles. This dataset is composed of two columns, with one containing the roles’ names and the other containing the corresponding paragraphs. For each target role, Mark1 and Tony, a binary classifier based on BERT was trained using the new dataset, with the target role labeled as ‘1’ and the other roles labeled as ‘0’. Therefore, the trained models can predict whether a paragraph belongs to the target role. The accuracy of the classifier for Mark1 on the validation set was 94\%, while the accuracy of the classifier for Tony was 96\%. In Task 1, each model generated a total of 10 conversations for each target role, with each conversation containing 10 paragraphs. Therefore, for each role-model pair, the corresponding classifier predicted 100 paragraphs. For each model, the total number of paragraphs predicted to belong to the target role was recorded. Additionally, a success rate was calculated for each model imitating each target role. It is equal to the number of paragraphs predicted to belong to the target role divided by the total number of paragraphs predicted. The results are shown in Table V. 

2) Evaluation for Task 2
To conduct the evaluation for Task 2, I used the dataset created during the human evaluation of Task 1 but replaced the paragraphs of Mark1 with 100 paragraphs of Mark2 from the test set of Dataset 3, as Mark2 is the target role for Task 2. Similar to Task 1, a binary classification model was trained using this updated dataset, and the accuracy on the validation set was 93\%. In Task 2, five conversations were generated by Llama 3 under each prompt, with each conversation containing 10 paragraphs. Therefore, for each prompt, a total of 50 paragraphs were predicted by the classifier. For each prompt, the total number of paragraphs predicted to belong to the target role was recorded. Similar to Task 1, a success rate was calculated for each prompt. The results are shown in Table VI.

\subsection{Results}

\subsubsection{ Results for Task 1}

From the results of the human evaluation, when LLMs were asked to imitate the language style of Mark1, Llama 3 obtained the highest score of 13.30, GPT-4 ranked second with a score of 12.83, and Gemini 1.5 received the lowest score of 11.87. For the performance of LLMs imitating language style of Tony, Llama 3 also performed the best with the score of 13.60, while GPT-4 achieved the second-highest score of 13.43, which is very close to Llama 3’s score. However, Gemini 1.5 received the lowest score, which is nearly 1 ponit less than the second-highest score. Therefore, based on the analysis of human evaluation for Task 1, it is evident that Llama 3 performed best at imitating the language styles of both Mark1 and Tony, while GPT-4 achieved the second-best performance for imitating these two roles. Gemini 1.5 demonstrated the lowest imitation capability, with scores significantly lower than those of the other two models. From the results of LLM evaluation, when imitating Mark1, Llama 3 performed the best with the highest score of 7.10, GPT-4 received the second highest score of 6.90, while Gemini 1.5 obtained the lowest score of 5.47. For imitating the language style of Tony, Llama 3 also perfomed the best with the score of 7.33, GPT-4 ranked the second with the score of 7.23, which is very close to Llama 3’s score, while Gemini 1.5 ranked the third with the score of 6.73. Based on the results of the LLM evaluation, the ranking of imitation capabilities, from most to least proficient, is as follows: Llama 3, GPT-4, Gemini 1.5.

From the results of the automatic evaluation, when the target role to be imitated was Mark1, Llama 3 achieved the highest success rate of 51.0\%. This indicates that 51.0\% of the paragraphs from conversations generated by Llama 3 were predicted by the classifier to belong to Mark1. GPT-4 achieved the second-higest success rate of 32\%, and Gemini 1.5 had the lowest success rate of 28\%. When imitating Tony, Llama 3 also had the highest success rate of 59\%, GPT-4 attained the second highest success rate of 48\%, while Gemini 1.5 had the lowest success rate of 29\%.

Synthesizing all evaluations from Task 1, it can be concluded that Llama 3 performs the best in imitating language styles when provided with a conversation and a designated role. GPT-4's imitation capacity, while slightly less robust, closely approximates that of Llama 3. In contrast, Gemini 1.5's imitation capability significantly lags behind that of the other models.

\subsubsection{ Results for Task 2}
For human evaluation, the ToT promting method got the highest score of 12.80, surpassing the scores of the other two prompts by over 1 point, The CoT prompt had the second higest score, while the standard prompt obtained the lowest score which is very close to that of the CoT prompt. Based on the these results, ToT method significantly improved the imitation ability of Llama 3, compared to the standard prompt and the CoT prompt. When looking at the results of LLM evaluation, ToT method also achieved the highest score of 6.10, while the standard prompt and the CoT prompt obtained scores of 4.60 and 4.80, respectively.

From the results of automatic evaluation, ToT prompting method had the highest success rate of 30.0\%, the CoT prompt obtained the second highest rate, 12\%, which is significantly lower than that of the ToT method. Standard prompt had the lowest success rate of 8\%, which is close to that of CoT method.

In summary, ToT prompting method greatly improved the Llama 3 model's ability to imitate language styles more effectively than the standard prompt and the CoT prompt. While the CoT prompt does offer a modest improvement over the standard prompt in imitation capabilities, the difference is relatively small.

\subsubsection{ Results for Task 3}

After executing the script developed for Task 3, a conversational AI based on the Llama 3 model and the ToT framework was successfully created. It’s a chatbot that can interact with users in the style of Mark2 when provided with Mark2 text in the prompt. An excerpt from my conversation with the chatbot about movies, shown in Fig. 5, demonstrates its ability to imitate Mark2's language style. The chatbot effectively mimics Mark2’s use of specific words and phrases, such as “you know” and “I mean”, and mirrors Mark2’s manner of speaking.

\subsubsection{ Conclusion}

This study compared the ability of three widely-used large language models to imitate the language style of a real person when provided with the same dialogue-form text and a zero- shot prompt. The results indicate that, under the same prompt, Llama 3 demonstrates the strongest imitation ability. GPT-4’s imitation ability is very close to that of Llama 3, while Gemini 1.5 significantly lags behind both of the former models. The study also examined how three different types of prompts affect the language style imitation ability of the same large language model, Llama 3. The results indicate that, compared to the standard zero-shot prompt, Llama 3’s ability to imitate language style is significantly improved with the ToT method. When guided by the chain-of-thought prompt, the model's imitation ability shows a slight improvement compared to the standard prompt, but the increase is minimal. Furthermore, this study designed a tree-of-thoughts framework and used it to guide Llama 3 in imitating the language style of a specific person. Therefore, a conversational AI that interacts with users in the language style of a real person has been created.

This study explores the ability of large language models to imitate the language styles based on only prompt guidance and a small amount of text. Different prompts are designed and their effects on a large language model are compared. This study provides a method for creating a conversational AI that mimics the language style of a real person. The method is based on the Llama 3 model and a Tree-of-Thoughts framework, achieving the imitation of a specific language style using only a small amount of text. This method eliminates the need for extensive text-based training, which can save costs. It also can be integrated as a key component into digital human technology and AI clone technology, allowing digital humans to possess a unique and authentic language style. However, this study is limited in scope, as it covers only three large language models and three different types of prompts. Future research will expand to evaluate more large language models and a wider variety of prompts, aiming to comprehensively explore the imitation capabilities of different models under various prompts. Additionally, it will continue to investigate the potential of combining large language models with prompt engineering techniques for enhancing language style imitation.

{\small
\bibliographystyle{ieee_fullname}
\bibliography{egbib}
}

\end{document}